\documentclass[journal,comsoc]{IEEEtran}
\usepackage[T1]{fontenc}

\ifCLASSINFOpdf
\else
\fi
\usepackage{amsmath}
\usepackage{cite}
\usepackage{amsmath,amsfonts}
\usepackage{amssymb}
\usepackage{dsfont}
\usepackage{amsthm}
\usepackage{algorithmic}
\usepackage{graphicx}
\usepackage{multicol}
\usepackage{textcomp}
\usepackage{eucal}
\usepackage[monochrome]{color}
\usepackage{xcolor}
\usepackage[normalem]{ulem}
\usepackage{xspace}
\usepackage{algorithm,algorithmic}
\newtheorem{ass}{Assumption}
\newtheorem{Lemma}{\rm \textbf{Lemma}}
\newtheorem{rem}{Remark}
\usepackage{xcolor}
\usepackage{etoolbox}

\makeatletter
\patchcmd{\@makecaption}
{\scshape}
{}
{}
{}
\makeatother

\newtheorem{innercustomcoro}{Corollary}
\newenvironment{customthcoro}[1]
  {\renewcommand\theinnercustomcoro{#1}\innercustomcoro}
  {\endinnercustomcoro}
  
\newtheorem{innercustomthm}{Theorem}
\newenvironment{customthm}[1]
  {\renewcommand\theinnercustomthm{#1}\innercustomthm}
  {\endinnercustomthm}
\usepackage[cmintegrals]{newtxmath}

\makeatletter
\newcommand\fs@betterruled{%
  \def\@fs@cfont{\bfseries}\let\@fs@capt\floatc@ruled
  \def\@fs@pre{\vspace*{5pt}\hrule height.8pt depth0pt \kern2pt}%
  \def\@fs@post{\kern2pt\hrule\relax}%
  \def\@fs@mid{\kern2pt\hrule\kern2pt}%
  \let\@fs@iftopcapt\iftrue}
\floatstyle{betterruled}
\restylefloat{algorithm}
\makeatother
\begin{document}
%
\title{Fast-Convergent Federated Learning with Adaptive
Weighting}
%
%
%

\author{Hongda Wu,~\IEEEmembership{}
        Ping Wang,~\IEEEmembership{Senior Member,~IEEE}
\thanks{Hongda Wu and Ping Wang are with the Department of Electrical Engineering and Computer Science, Lassonde School of Engineering, York University, Toronto, ON M3J 1P3, Canada (e-mail: hwu1226@eecs.yorku.ca; pingw@yorku.ca)}
\thanks{}
}
\maketitle

\begin{abstract}
Federated learning (FL) enables resource-constrained edge nodes to collaboratively learn a global model under the orchestration of a central server while keeping privacy-sensitive data locally. The non-independent-and-identically-distributed (non-IID) data samples across participating nodes slow model training and impose additional communication rounds for FL to converge. In this paper, we propose \texttt{Fed}erated \texttt{Ad}a\texttt{p}tive Weighting (\texttt{FedAdp}) algorithm that aims to accelerate model convergence under the presence of nodes with non-IID dataset. We observe the implicit connection between the node contribution to the global model aggregation and data distribution on the local node through theoretical and empirical analysis. We then propose to assign different weights for updating the global model based on node contribution adaptively through each training round. The contribution of participating nodes is first measured by the angle between the local gradient vector and the global gradient vector, and then, weight is quantified by a designed non-linear mapping function subsequently. The simple yet effective strategy can reinforce positive (suppress negative) node contribution dynamically, resulting in communication round reduction drastically. Its superiority over the commonly adopted Federated Averaging (\texttt{FedAvg}) is verified both theoretically and experimentally. With extensive experiments performed in Pytorch and PySyft, we show that FL training with \texttt{FedAdp} can reduce the number of communication rounds by up to 54.1\% on MNIST dataset and up to 45.4\% on FashionMNIST dataset, as compared to \texttt{FedAvg} algorithm.
\end{abstract}

\begin{IEEEkeywords}
Federated Learning, Communication Effeciency, Mobile Edge Computing, Internet of Things.
\end{IEEEkeywords}

%
\IEEEpeerreviewmaketitle

\section{Introduction}
%
%
%
%
\IEEEPARstart{T}{he} rapid advancement of edge devices (e.g., Internet of Things (IoT), mobile phones) is constantly generating an unprecedented amount of data \cite{Knud2019}. These devices are currently equipped with enhanced sensors, computing, and communication capability. Coupled with the rise of Deep Learning (DL) \cite{lecun2015deep}, the edge devices unfold the countless opportunities for various tasks of modern society, e.g., road congestion prediction \cite{zhang2015testing} and environmental monitoring \cite{ganti2011mobile}.

In the traditional cloud-centric approaches, data generated and collected by edge devices is uploaded and processed in a data center. It is predicted that the data generation rate will exceed the capacity of today's Internet in the near future \cite{chiang2016fog}, Mobile Edge Computing (MEC) has naturally been proposed to incorporate the data processing outside the cloud \cite{xiong2018mobile,wang2020convergence}. With computing and storage capability, MEC systems generally consist of end-edge-server architecture. Multiple edge servers are capable of performing large-scale distributed tasks involving local processing and remote execution under the coordination of a remote cloud. MEC approaches compromise training efficiency and communication cost by bringing model training towards where the data is generated. However, computation offloading task and data processing at the edge server still involves the transmission of sensitive data.

In either centralized cloud training or MEC approaches, collecting data for model training is unrealistic from a privacy, security, regulatory, or necessity perspective. In order to maintain privacy-sensitive data and to facilitate collaborative machine learning (ML) among distributed nodes, Federated Learning (FL) has emerged as an attractive paradigm, where local nodes collaboratively train a task model under the orchestration of a central server without accessing end-user data \cite{McMahan2017CommunicationEfficientLO, konevcny2016federated}. In FL, local nodes cooperatively train an ML \textcolor{blue}{model} required by the central server by utilizing their local data. Through transferring local model updates to the central server for model aggregation and acquiring a global model for local training rather than sending raw data,  user data privacy is well protected. As such, FL features from conventional approaches in data acquisition, storage, and training. FL has been deployed by major service providers and plays an important role in supporting privacy-sensitive applications, including computer vision, natural language processing, and medical database\cite{lim2020federated}.

Even though good convergence performance of FL approach is shown, owing to limited connectivity of wireless networks, the availability of local nodes and straggler of participating nodes, communication cost becomes a critical bottleneck in FL context since generally several iterations are involved for model converging\cite{McMahan2017CommunicationEfficientLO, konevcny2016federated, lim2020federated}. Another fundamental challenge for FL is strongly non-independent-and-identically-distributed (non-IID) and highly skewed data across local nodes. The presence of non-IID data significantly degrades the performance of federated learning, which makes model training take more rounds to converge, and the variance caused by non-IID data brings instability to the training process \cite{zhao2018federated, wang2020optimizing, li2018federated}. Since the completion time of federated learning is largely impacted by the communication time, how to reduce the communication round for model convergence in FL, especially for participating nodes with non-IID datasets, is urgent to be addressed.

In this paper, to surmount the slow convergence of vanilla Federated Averaging (\texttt{FedAvg}) \cite{McMahan2017CommunicationEfficientLO} under the presence of non-IID dataset, we propose \texttt{Fed}erated \texttt{Ad}a\texttt{p}tive Weighting (\texttt{FedAdp}) algorithm that aims to improve the performance of federated learning through assigning distinct weight for participating node to update the global model. We observe that nodes with heterogeneous datasets make different contributions to the global model aggregation. Therefore, our main intuition is to measure the contribution of the participating node based on the gradient information from local nodes then assign different weights accordingly and adaptively at each communication round for global model aggregation. According to node contribution, the proposed adaptive weighting strategy is capable of reducing the expected training loss of FL in each communication round under the presence of non-IID nodes, which accelerates the model convergence.
Our main contributions in this
paper are as follows:
\begin{itemize}
\item We identify the presence of nodes with non-independent-and-identically-distributed (non-IID) data distributions slows the convergence speed of federated learning. In addition, we analyze the convergence bound of gradient-descent based federated learning from a theoretical perspective and derive the convergence bound that incorporates the non-IID data distribution across participating nodes and weighting strategy for model updating.
\item We observe the implicit connection between data distribution on a node and the contribution from that node to the global model aggregation, measured at the central server-side by inferring gradient information of participating nodes. The convergence bound is lowered, and the convergence speed is accelerated by a carefully designed weighting strategy, which is formalized as \texttt{Fed}erated \texttt{Ad}a\texttt{p}tive Weighting (\texttt{FedAdp}), that assigns different weights to nodes for global model aggregation in each round of communication.
\item We empirically evaluate the performance of the proposed weighting algorithm via extensive experiments using different real datasets with different learning objectives (i.e., convex and non-convex loss function). Our experimental results have shown that FL training with \texttt{FedAdp} can drastically reduce the communication rounds compared with the commonly adopted \texttt{FedAvg} algorithm.
\end{itemize}

The rest of this paper is organized as follows. Section  \ref{II} discusses the related works. Section \ref{III} provides the preliminaries of federated learning and the impact of non-IID data on FL learning. In Section \ref{IV}, the convergence analysis and the proposed weighting algorithm are presented. Experimental results are shown in Section \ref{V} and the conclusion is
presented in Section \ref{VI}.

\section{Related Work}
\label{II}
Generally, the FL algorithm adopts synchronous aggregation and selects a subset of nodes randomly to participate in each round randomly to avoid long-tailed waiting time due to the network uncertainty and straggler. To boost convergence and reduce the communication rounds, tuning the number of local updates\cite{McMahan2017CommunicationEfficientLO, wang2019adaptive, tran2019federated, li2018federated}, and selecting appropriate nodes for FL training \cite{nishio2019client, wang2020optimizing, nguyen2020fastconvergent} are the usually adopted approaches. 

In particular, McMahan \textit{et al.}\cite{McMahan2017CommunicationEfficientLO} presented the vanilla Federated Averaging (\texttt{FedAvg}) algorithm, which increases the number of local updates instead of updating the local model one time at each round. Li \textit{et al.} \cite{li2018federated} proposed to allow participating nodes to perform a variable number of local updates, rather than applying the same amount of workload for each node \cite{McMahan2017CommunicationEfficientLO}, to consequently overcome the heterogeneity of the system. Similar to \cite{li2018federated}, authors in \cite{tran2019federated} also posed local accuracy for participating nodes, based on limited computing resources on nodes, as an index to steer the number of local updates performed. Different from \cite{li2018federated, tran2019federated}, the work in \cite{wang2019adaptive} exposed an analytical model to dynamically adapt the number of local updates between two consecutive global aggregations in real-time to minimize the learning loss under a fixed resource budget of the edge computing system. Regarding the node selection,
Nishio \textit{et al.} \cite{nishio2019client} proposed \texttt{FedCS} algorithm to do node selection intentionally rather than randomly, based on the resource conditions of local nodes. Authors in \cite{nguyen2020fastconvergent} utilized gradient information to do node selection. The nodes whose inner product between its gradient vector and the global gradient vector is negative will be excluded from FL training. 

To handle the non-IID data distribution,
Zhao \textit{et al.} \cite{zhao2018federated} quantified the weight divergence by earth mover's distance between data distribution on nodes and population distribution. However, the strategy of pushing a small set of uniformly distributed data to participating nodes in \cite{zhao2018federated} violates the privacy concern of FL and imposes extra communication cost. It was proposed in \cite{wang2020optimizing} that communication rounds can be reduced effectively by selecting nodes based on their uploaded model weights, which profile the data distribution on those nodes. In contrast, Wang \textit{et al.}\cite{8885054} proposed to identify the irrelevant update caused by different data distribution at the node side. 
The communication cost is accordingly reduced by precluding these nodes with irrelevant updates before updates transmission. However, local nodes are required to check the relevance in each round using the global model kept in the previous round, which is in contravention of FL and brings computational burdens to local nodes.

Regarding the weighting strategy, authors in \cite{chen2019communication} proposed to assign different weights for global model aggregation adaptively by considering the time difference when the model update is done in a layerwise asynchronous manner.  Chai \textit{et al.}\cite{chai2020fedat} designed a tier-based FL system by dividing the participating nodes into tiers according to their responding time and devised to adaptively assign weights to different tiers for model aggregation since there exists different updating frequency across tiers. Both methods in \cite{chen2019communication, chai2020fedat} aim to weigh the local update along with different communication rounds.

To enhance the convergence of FL with the presence of non-IID nodes, different from \cite{zhao2018federated, wang2020optimizing} that measure model weight, we find out that nodes contribute differently to the global model aggregation owing to their different data distribution, and there exists an implicit connection between data distribution and gradient information. In this paper, we propose to measure the node contribution quantitatively by the angle between the local gradient at the nodes and the global gradient across all participating nodes at the server-side. With the quantified contribution, the weight for aggregating the global model can be devised discriminatively across the nodes and adaptively in each round according to node contribution. The proposed adaptive weighting strategy can effectively speed up the convergence of FL in the presence of non-IID data. Different from \cite{nguyen2020fastconvergent, 8885054}, our method does not impose additional communication and computation burden to local nodes. Besides, our adaptive weighting strategy is done in each communication round, which is orthogonal with the methods proposed in \cite{chen2019communication, chai2020fedat}.

\section{Preliminaries}
\label{III}
In this section, we briefly introduce key ingredients behind the recent method for federated learning, \texttt{FedAvg}, and show how non-IID data impacts model convergence.

\subsection{Standard Federated Learning }
\label{III-A}
In general, federated learning methods \cite{McMahan2017CommunicationEfficientLO}\cite{lim2020federated} are designed to handle the consensus learning task in a decentralized manner, where a central server coordinates the global learning objective and multiple devices training the local model with locally collected data. In particular, assume that we have $N$ local nodes with dataset $\mathcal{D}_1,..., \mathcal{D}_i, ..., \mathcal{D}_N$ and we define $D_i \triangleq |\mathcal{D}_i|$ \textcolor{blue}{as the number of data samples owned by each node}, where $| \cdot |$ denotes \textcolor{blue}{the Cardinality of sets.} FL methods aim to minimize:
\begin{align*}
\underset{\mathbf{w}}{\min} \quad F( \mathbf{w} )  \triangleq \sum _{i=1}^{N}\psi_{i} F_{i}(\mathbf {w}), \tag{1}
\end{align*}
where $\mathbf{w}$ is global model weight, \textcolor{blue}{$\psi_i = D_i/ \sum_{i^\prime=1}^{N} D_{i^\prime}$} is the weight for aggregation in FL training, and global objective function $F( \mathbf{w} )$ is surrogated by using local objective function $F_i( \mathbf{w} )$, which is defined, as an example, in the context of $C$-class classification problem thereinafter. In particular, $C$-class classification problem is defined over a feature space $\mathcal{X}$ and a label space $\mathcal{Y} = [C]$, where $[C] = \{1, \cdots, C \}$. For each labeled data sample $\{\mathbf {x}, y \}$, predicted probability vector $\widetilde {\mathbf {y}}$ is achieved by using mapping function $f: \mathcal{X} \rightarrow \widetilde{\mathcal{Y}}$, where $\widetilde{\mathcal{Y}} = \{\widetilde {\mathbf {y}}|\sum_{j=1}^C \widetilde y_j =1, \widetilde y_j \geq 0, \forall j \in [C]   \}$. As such, $F_i( \mathbf{w} )$ commonly measures the local empirical risk over possibly differing data distribution $p^{(i)}$ of node $i$, which is defined by using cross entropy for $C$-class classification as follow,
\begin{align*}
\underset{\mathbf{w}}{\min} \quad F_i( \mathbf{w} ) & \triangleq 
\mathbb{E}_{\mathbf{x},y \backsim p^{(i)}} \left[- \sum_{j=1}^C {\mathds{1}}_{y=j} {\rm log}f_j (\mathbf{x},\mathbf{w})  \right] \\
& = - \sum_{j=1}^C p^{(i)} (y=j) \mathbb{E}_{\mathbf{x}|y=j} \left[ {\rm log}f_j (\mathbf{x},\mathbf{w}) \right], \tag{2}
\end{align*}
where $f_j (\mathbf{x},\mathbf{w})$ denotes the probability that the data sample $\mathbf{x}$ is classified as the $j$-th class given model $\mathbf{w}$, and $p^{(i)} (y=j)$ denotes the data distribution on node $i$ over class $j \in [C]$.

In \textcolor{blue}{general federated learning setting (e.g., \texttt{FedAvg})}, the participating nodes perform local training with the same training configuration (e.g. optimizer, learning rate, etc).
At each communication round $t$, a subset of the nodes $\mathcal{S}_t, |\mathcal{S}_t| = K \ll N$ are selected and global model $\mathbf{w} ({t-1})$ in previous iteration is sent to the selected nodes. Each of the participating nodes $i$ performs stochastic gradient descent (SGD) training to optimize its local objective $F_i( \mathbf{w})$: 

\begin{align*}
\label{eq3}  
 \mathbf{w}_i (t)  =  \mathbf{w} (t-1) - \eta \nabla F_i(\mathbf{w} (t-1)) 
 ,\tag{3}
\end{align*}
where $\eta$ is the learning rate and $\nabla  F_i(\cdot)$ is the gradient at node $i$. \textcolor{blue}{(\ref{eq3}) gives a general principle of SGD optimization.} $ \mathbf{w}_i (t)$ could be the result after one or several local updates of SGD \textcolor{blue}{  (e.g., $\tau =1$ in \texttt{FedSGD}\cite{McMahan2017CommunicationEfficientLO} or $\tau>1$ in \texttt{FedAvg}\cite{McMahan2017CommunicationEfficientLO, wang2019adaptive} with $\tau$ denoting the number of local updates between two consecutive global rounds). Hereinafter, SGD is applied to mini-batch data samples with size $\bar B$. As such, local model is updated by $\tau = 
\frac{D_i}{\bar B}E$ times, where $D_i$ and $E$ are the number of training samples on node $i$ and the number of local training epochs, respectively.}

The nodes then communicate their local model updates $\Delta_i(t) = \mathbf{w}_i (t)	- \mathbf{w} (t-1) $ to the central server\footnote{Typically there are two ways for nodes to upload their local model to the
central server, either by uploading model parameters $\mathbf{w} (t)$ or by uploading
the model difference $\Delta_i(t)$. Although the same amount of data are to be uploaded in both ways, conveying $\Delta_i(t)$ is proven to be more amenable for compression \cite{konevcny2016federated}.}, which aggregates them and updates the global model accordingly,
\begin{align*}
\label{eq4}
 \Delta (t) &=  \sum _{i=1}^{|\mathcal{S}_t|}\psi_{i}\Delta_i(t) \\
 \mathbf{w} (t) &= \mathbf{w} (t-1) +  \Delta (t).
  \tag{4}
\end{align*}

\subsection{{\rm\texttt{FedAvg}} for non-IID data}
\label{III-B}
The independent and identically distributed (IID) sampling condition of training data is important that the stochastic gradient is an unbiased estimate of the full gradient \cite{wang2019adaptive}. \texttt{FedAvg} is shown to be effective, given that the data distribution across different nodes is the same as centrally collected data. However, the data distribution determined by usage patterns across local nodes is typically non-IID, i.e., $p^{(i)}$ is different across participating nodes. 

Since local objective $F_i( \mathbf{w} )$ is closely related with data distribution $p^{(i)}$, a large number of local updates lead the model towards optima of its local objective $F_i( \mathbf{w} )$ as opposed to the global objective $F( \mathbf{w} )$. The inconsistency between local models $ \mathbf{w}_i$ and global model $ \mathbf{w}$ is accumulated along with local training, leading to more communication rounds before training converges. As such, local training with multiple local updates potentially hurts convergence and even leads to divergence with the presence of non-IID data \cite{McMahan2017CommunicationEfficientLO}\cite{zhao2018federated}. 

We conduct an experiment to demonstrate the impact of non-IID data on model convergence. We train a two-layer CNN model with the same neural network architecture in \cite{McMahan2017CommunicationEfficientLO} using Pytorch on the MNIST dataset (containing 60,000 samples with 10 classes) until the model achieves 95\% test accuracy. 10 nodes are selected,  each with 600 samples that are selected based on their label criteria. If a node is at \textit{IID setting}, 600 samples are randomly selected over the whole training set. If a node is at \textit{$x$-class non-IID setting}, 600 samples are randomly selected over a subset, which is composed of $x$ class data samples. Each class of the $x$-class is selected at random and can be overlapped. The skewness of datasets is measured and reflected by the value of $x$.

\begin{figure}[h]
\centerline{\includegraphics[scale=0.55]{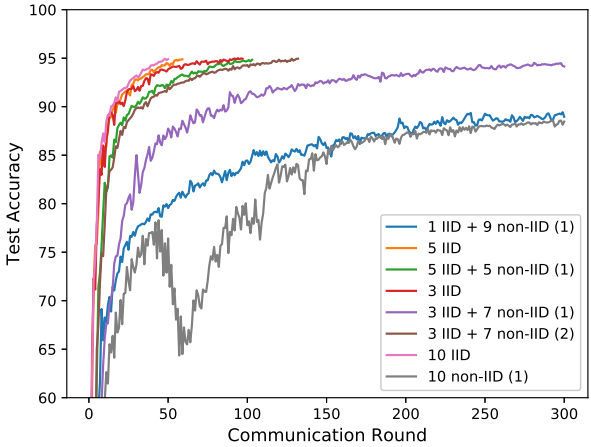}}
\caption{Test accuracy over communication rounds of \texttt{FedAvg} with heterogeneous data distribution over participating nodes. $\rm X$ IID + $\rm Y$ non-IID (1) (or (2)) represents $\rm X$ nodes are at \textit{IID setting} and $\rm Y$ nodes are at \textit{1-class (or 2-class) non-IID setting}}
\label{fig1}
\end{figure}

We use the same notations for \texttt{FedAvg} algorithm as \cite{McMahan2017CommunicationEfficientLO}: $\bar B$, the local minibatch size, and $E$, the number of local \textcolor{blue}{training} epochs. In this experiment, $\bar B=32$, $E=1$, $\eta=0.01$ and learning rate decay of $0.995$ per communication round. We can conclude from Fig. \ref{fig1}:
\begin{itemize}
\item Model convergence highly depends on IID nodes. The presence of non-IID nodes imposes variance to model training, which slows the convergence of FL (e.g., 5 IID case converges faster than  5 IID + 5 non-IID (1) case).
\item The skewness of data affects model convergence. With the participation of the non-IID node, the model converges much slower when the skewness of the dataset increases (e.g., 3 IID + 7 non-IID (2) case converges much faster than 3 IID + 7 non-IID (1) case).
\end{itemize}

\section{Federated Adaptive Weighting}
\label{IV}
In this section, we develop our methodology for improving the convergence rate of federated learning. We first analyze the convergence property of federated learning (Section \ref{IV-A}). The theoretical analysis on the expected decrease of FL loss in each round of training reveals that gradient information and data distribution impact the convergence. The experimental result shows the diversity of node contribution in reducing the FL loss in each round (Section \ref{IV-B}), measured by the local gradient of each node and the global gradient from participating nodes. This motivates us to assign weight adaptively according to node contribution for global model aggregation.  Finally, we theoretically prove that assigning weight based on node contribution adaptively leads to accelerating model convergence and formally present the methodology of the proposed \texttt{FedAdp} algorithm (Section \ref{IV-C}).

\subsection{Convergence Analysis}
\label{IV-A}
For theoretical analysis of federated learning algorithms, we employ the following typical assumptions in our analysis (see e.g., \cite{zhao2018federated, wang2019adaptive, li2018federated, nguyen2020fastconvergent}).
\begin{ass} 
\label{ass1}
\textbf{$\beta$-Lipschitz smoothness}.\\
$ F_i( \mathbf{w})$ is $\beta$-Lipschitz smoothness for each of the participating nodes $i \in \mathcal{S}_t$, 
 i.e., $\Vert \nabla F_i( \mathbf{w}) -\nabla F_i( \mathbf{w'}) \Vert  \leq \beta \Vert  \mathbf{w} -  \mathbf{w'} \Vert $ for any  two parameter vectors $ \mathbf{w}$, $ \mathbf{w'}$.
\end{ass}
Based on Assumption \ref{ass1}, the definition of $F( \mathbf{w}) $, and  triangle inequality, we can easily get the following lemma.
\begin{Lemma}
\label{lem1}
\textit{$F( \mathbf{w}) $ is $\beta$-Lipschitz smoothness.}
\end{Lemma}

\begin{ass} 
\label{ass2}
\textbf{Bounded Local Dissimilarity \footnote{ Similar assumption has made in FL context, for example in \cite{wang2019adaptive, li2018federated, nguyen2020fastconvergent}. In\cite{li2018federated, nguyen2020fastconvergent}, the dissimilarity across local gradients is imposed by an upper bound
to capture the impact of data heterogeneity on FL convergence, and an analogous definition named gradient divergence is also presented in \cite{wang2019adaptive}. 
By tracking the divergence of gradients on each participating node, we observe that the dissimilarity can be further bounded by a lower bound as shown in Assumption \ref{ass2}.}} \\
For any participating node $i$, the dissimilarity between local objective and global objective at $\mathbf{w}$ is bounded by $A$ and $B$, i.e., $ A \Vert \nabla F( \mathbf{w}) \Vert \leq  \Vert \nabla F_i( \mathbf{w}) \Vert \leq B \Vert  \nabla  F( \mathbf{w}) \Vert$.
\end{ass}

Here $\nabla F( \mathbf{w})$ is the gradient of the global objective that is defined as $ \nabla F( \mathbf{w}) = \sum_{i=1}^{|\mathcal{S}|} \textcolor{blue}{(D_i / \sum_{i^\prime=1}^{|\mathcal{S}|} D_{i^\prime})} \nabla F_i( \mathbf{w}) $ in FL context. The local dissimilarity in assumption \ref{ass2} can be seen as a metric that reveals the data heterogeneity when the same training configuration (e.g., learning rate, batch size, training epoch, etc.) across participating nodes is held. As a sanity check, when all the local data samples are the same, we have $A = B = 1$.

\begin{customthm}{1}
\label{thm1}
With loss function $F_i( \mathbf{w})$ satisfying Assumptions \ref{ass1}-\ref{ass2} and supposing $\mathbf{w}(t)$ is not a stationary solution, the expected decrease in the global loss function between two consecutive rounds satisfies,
\begin{align*}
\label{eq5}
& F(\mathbf{w}(t+1)) \leq  F( \mathbf{w}(t))  \\ & -\eta \mathbb{E}_{i|t} \left[ \left(  \frac{\langle \nabla  F( \mathbf{w}(t)), \nabla  F_i( \mathbf{w}(t))\rangle}{\Vert \nabla  F( \mathbf{w}(t))\Vert  \Vert \nabla  F_i( \mathbf{w}(t))\Vert} - \frac{B \beta\eta}{2} \right) \cdot \frac{A^2}{B} \Vert \nabla F( \mathbf{w}) \Vert ^2 \right], \tag{5}
\end{align*}
where the expectation $\mathbb{E}_{i|t}$ refers to the weighting strategy of the participating node $i \in \mathcal{S}_t$ for global model aggregation. $\langle \cdot \rangle$ is the inner product operation and $\Vert \cdot \Vert$ denotes the $\ell2$ norm of a vector.
\end{customthm}
The proof of Theorem \ref{thm1} is presented in Appendix\ref{apex1}. Theorem \ref{thm1} provides a bound on how rapid the decrease of the global FL loss can be expected. Based on Theorem \ref{thm1}, we have the following corollary and remarks.

\begin{customthcoro}{1}
\rm The convergence upper bound of FL after $T$ global rounds is given by,
\begin{align*}
\label{6}
& F(\mathbf{w}(T)) \leq  F( \mathbf{w}(0))  \\ & -\eta \sum_{t=0}^{T-1}   \mathbb{E}_{i|t} \left[ \left(  \frac{\langle \nabla  F( \mathbf{w}(t)), \nabla  F_i( \mathbf{w}(t))\rangle}{\Vert \nabla  F( \mathbf{w}(t))\Vert  \Vert \nabla  F_i( \mathbf{w}(t))\Vert} - \frac{B\beta\eta}{2} \right) \cdot \frac{A^2}{B} \Vert \nabla F( \mathbf{w}) \Vert ^2 \right]. \tag{6}
\end{align*}
\end{customthcoro}
\begin{rem}
\label{rem1}
\rm
The decrease of FL loss between two consecutive global rounds shows a dependency on learning rate $\eta$, the bounded local dissimilarity of participating nodes, the correlation between the local gradient and the global gradient $\frac{\langle \nabla  F( \mathbf{w}(t)), \nabla  F_i( \mathbf{w}(t))\rangle}{\Vert \nabla  F( \mathbf{w}(t))\Vert  \Vert \nabla  F_i( \mathbf{w}(t))\Vert}$, and the weight strategy $\mathbb{E}_{i|t}$ \textcolor{blue}{that weighs} participating nodes for the global model aggregation in each global round.
\end{rem}

\begin{rem}
\label{rem2}
\rm
The local gradient, which is correlated with minimizing the local objective, may not align with the direction of approaching the optimal of the global objective. The correlation $\frac{\langle \nabla  F( \mathbf{w}(t)), \nabla  F_i( \mathbf{w}(t))\rangle}{\Vert \nabla  F( \mathbf{w}(t))\Vert  \Vert \nabla  F_i( \mathbf{w}(t))\Vert}$ between the local gradient and the global gradient is a metric to measure their alignment level. From Theorem \ref{thm1}, we can see this metric also indicates how much each node contributes to reducing FL loss in each global round.
\end{rem}

\begin{rem}
\label{rem3}
\rm The FL loss $F(\mathbf{w}(t+1))$ is negatively associated with the bound gap imposed in Assumption \ref{ass2}, meaning that as bound gap $[A,B]$ grows larger, the bound weakens, and the convergence exacerbates. Intuitively, the root cause of dissimilarity is the divergence of local gradients across participating nodes with heterogeneous datasets, which can be intentionally regularized by a properly designed weighting strategy.
\end{rem}

An immediate suggestion from Theorem \ref{thm1} that to improve the convergence of FL, one can reduce the FL loss by \textcolor{blue}{increasing} $\mathbb{E}_{i|t} \left[ \cdot \right]$ in each global round. This motivates us to measure node contribution quantitatively through the correlation $\frac{\langle \nabla  F( \mathbf{w}(t)), \nabla  F_i( \mathbf{w}(t))\rangle}{\Vert \nabla  F( \mathbf{w}(t))\Vert  \Vert \nabla  F_i( \mathbf{w}(t))\Vert} $ between the local gradient and the global gradient and assign larger weights  \textcolor{blue}{to the nodes with} higher contribution to \textcolor{blue}{enlarge} the expected \textcolor{blue}{decrease of} FL loss in each global round.
\subsection{Measurement of Node Contribution}
\label{IV-B}
In FL, the direction of minimizing local objective $F_i(\mathbf{w})$ might not align with the direction of minimizing $F(\mathbf{w})$. In particular, it can be deduced from (\ref{eq3}) that the gradient on different nodes may be tremendously diverse, especially for heterogeneous datasets across participating nodes. As such, the contribution from participating nodes for global aggregation is different. From our experiment, we note that if the data distribution on a node is highly skewed, the gradient may highly deviate from or even in the opposite direction to the global gradient, causing a negative effect on the global aggregation.

Instead of assigning weight for participating nodes based on the size of datasets as in \texttt{FedAvg} \cite{McMahan2017CommunicationEfficientLO}, we measure the contribution of participating nodes based on the correlation between local gradient and global gradient. Particularly, we quantify the contribution of each node at each global round based on \textit{angle $\theta_i(t)$}, that is defined as:
\begin{align*}
\label{eq8}
\theta_i(t) & = \arccos{ \frac{\langle\  \nabla F( \mathbf{w}(t)) , \nabla  F_i( \mathbf{w}(t))  \rangle}{\Vert \nabla F( \mathbf{w}(t))   \Vert  \Vert \nabla F_i( \mathbf{w}(t)) \Vert} }. \tag{8}
\end{align*}

From (\ref{eq8}), we can see that when the angle $\theta_i(t)$ is small, it means the local gradient $\nabla F_i( \mathbf{w}(t))$ has a similar direction to the global gradient, thereby positively contributing to the global aggregation. In contrast, when $\theta_i(t)$ is large, e.g., larger than $\pi/2$, the local gradient $\nabla F_i( \mathbf{w}(t))$ has an opposite direction to the global gradient, thereby negatively contributing to the global aggregation.

To restrain the instability caused by randomness presented in  instantaneous angle $\theta_i(t) $ at each round, we use so-called \textit{smoothed angle $\widetilde \theta_i(t)$} as a substitution, which is the averaged angle over previous training rounds and is defined as:
\begin{align*}
\label{eq9}
\widetilde \theta_i(t) =
    \begin{cases}
      \theta_i(t) & t=1\\          
  \frac{t-1}{t}\widetilde \theta_i(t-1) + \frac{1}{t} \theta_i(t) & t>1
    \end{cases}  
 .    \tag{\textcolor{blue}{9}}
\end{align*}

By using \textit{smoothed angle $\widetilde \theta_i(t)$}, the angle difference across nodes uniquely depends on the data distribution. Intuitively, the angle $\widetilde \theta_i(t)$ will be larger as the dissimilarity between data distribution on node $i$ and population distribution grows. Also, the smoothed angle is capable of quantifying the degree of data dissimilarity among the local nodes. 

\begin{figure}[t]
\centerline{\includegraphics[scale=0.55]{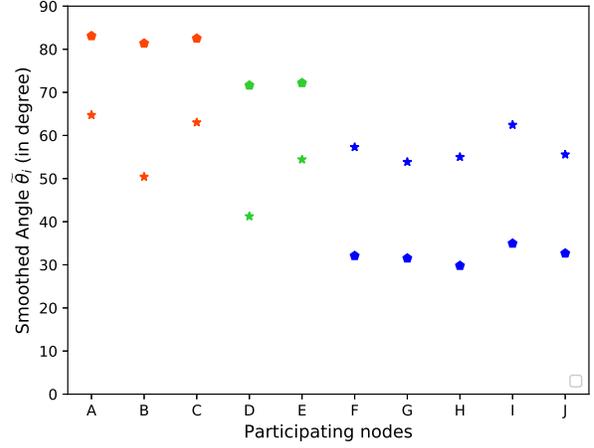}}
\caption{The smoothed angle $\widetilde \theta_i$ of participating node at different training round, where star and pentagon sign denote the angle at commmunication round 1 \textcolor{blue}{and} communication round 15, respectively. Nodes with different data distribution are marked with different colors.}
\label{fig2}
\end{figure}
We conduct an experiment to illustrate how data distribution can be reflected by angle. Under the same training model in \ref{III-B}, we randomly assign i) 3 nodes with \textit{$1$-class non-IID setting} \textcolor{blue}{(i.e., node $\rm ``A", ``B", ``C"$)}, ii) 2 nodes with \textit{$2$-class non-IID setting} \textcolor{blue}{(i.e., nodes $\rm ``D"$ and $``\rm E"$)}, and iii) the rest of 5 nodes with \textit{IID setting}. 

As shown in Fig. \ref{fig2}, the smoothed angle between the local gradient and the global gradient is full of randomness at the beginning of FL training. Along with the training, smoothed angle $\widetilde \theta_i$ shows diversity across the participating nodes due to the impact of data heterogeneity on local training. \textcolor{blue}{To be more specific, for those nodes with $1$-class non-IID setting, the data samples from which are highly skewed since the label space $\mathcal{Y}$ is extremely limited.  Due to the limited richness of data samples on node $i$, the direction for minimizing its local objective $F_i(\mathbf{w})$, which is reflected by $\nabla F_i(\mathbf{w})$, will be far away from the direction for minimizing the overall objective $F(\mathbf{w})$, which is reflected by  $ \nabla F( \mathbf{w}) = \sum_{i=1}^{|\mathcal{S}|} (D_i / \textcolor{blue}{\sum_{i^\prime=1}^{|\mathcal{S}|} D_{i^\prime}}) \nabla F_i( \mathbf{w}) $, resulting a greater $\theta_i$ as defined by (\ref{eq8}). As shown in Fig. \ref{fig2}}, the gradient from the node with extremely skewed data (e.g., node $\rm ``A", ``B", ``C"$) is nearly orthogonal with the global gradient after 15 communication rounds, which barely brings a contribution to the global model. If we ignore the discrepancy of node contribution and average local update according to the size of datasets, as in \texttt{FedAvg}, it slows model convergence.

\subsection{\texttt{Fed}erated \texttt{Ad}a\texttt{p}tive Weighting \rm (\texttt{FedAdp})}
\label{IV-C}
Provided the diverse node contribution from participating nodes, the weighting strategy affects Theorem \ref{thm1} through the expectation $\mathbb{E}_{i|t} \left[ \frac{\langle \nabla  F( \mathbf{w}(t)), \nabla  F_i( \mathbf{w}(t))\rangle}{\Vert \nabla  F( \mathbf{w}(t))\Vert  \Vert \nabla  F_i( \mathbf{w}(t))\Vert} \right]$ consequently. To \textcolor{blue}{accelerate} the convergence rate, we seek to \textcolor{blue}{lower} the upper bound of the expected loss in each communication round, which reveals to assign different weights $\widetilde  \psi_i$ to different nodes for the global model aggregation. As such, the corresponding \textcolor{blue}{objective} is formally stated as \textcolor{blue}{enlarging $\mathbb{E}_{i|t} \left[ \frac{\langle \nabla  F( \mathbf{w}(t)), \nabla  F_i( \mathbf{w}(t))\rangle}{\Vert \nabla  F( \mathbf{w}(t))\Vert  \Vert \nabla  F_i( \mathbf{w}(t))\Vert} \right] =  \sum_{i}^{|\mathcal{S}_t|} \frac{\langle \nabla  F( \mathbf{w}(t)), \nabla  F_i( \mathbf{w}(t))\rangle}{\Vert \nabla  F( \mathbf{w}(t))\Vert  \Vert \nabla  F_i( \mathbf{w}(t))\Vert} \cdot  \widetilde \psi_i(t) $} via designing $\widetilde  \psi_i$ \textcolor{blue}{under the inherent constrain $\sum_{i}^{|\mathcal{S}_t|} \widetilde  \psi_i(t)  =1, \quad  \widetilde \psi_i(t)  \geq 0 \; \; \forall i, t$}.

Considering the node contribution is measured by (\ref{eq8}), 
a natural weighting design aiming to  \textcolor{blue}{enlarge}  the expectation should follow the criterion that nodes with higher contribution deserve higher weights for aggregation in each global round. We characterize the contribution-regulated weighting strategy for the global aggregation in each global round adaptively as \texttt{Fed}erated \texttt{Ad}a\texttt{p}tive Weighting \rm (\texttt{FedAdp}).

Assigning adaptive weight for updating the global model in the proposed \texttt{FedAdp} algorithm includes two steps:

\subsubsection{Non-linear mapping function}
\label{IV-C-1}
We design a non-linear mapping function to first quantify the contribution of each node based on angle information. Inspired by \textcolor{blue}{the} sigmoid function, we use a variant of \textit{Gompertz function}\cite{883477}, which is a non-linear decreasing function defined as
 \begin{align*}
\label{eq13}
f(\widetilde \theta_i(t)) = \alpha (1 - e^{-e^{-\alpha(\widetilde \theta_i(t)-1)}}),  \tag{10}
\end{align*}
where $\widetilde \theta_i(t) $ is the \textit{smoothed angle} in \textit{radian}, $e$ denotes the exponential constant and $\alpha$ is a constant as explained in the following.

The designed mapping function has several properties that are important for the subsequent weight calculation:
\begin{itemize}
\item $\lim_{\widetilde \theta_i(t) \to \pi/2} f(\widetilde \theta_i(t)) =\epsilon$, where $\epsilon \varpropto \frac{1}{\alpha} $ is constant; 
\item $\lim_{0 \to \widetilde \theta_i(t) \to \upsilon } f(\widetilde \theta_i(t)) =\alpha $, where $\upsilon \varpropto \alpha$ is a constant;
\end{itemize}

$\alpha$ controls the decreasing rate \textcolor{blue}{of $f(\widetilde \theta_i(t))$} from $\alpha$ to $\epsilon$ as $\widetilde \theta_i(t)$ increases from $\upsilon$ to  $\pi/2$. \textcolor{blue}{For example, a small $\alpha \in \mathbb{Z}^{+}$ indicates a lower decreasing rate of $f(\widetilde \theta_i(t))$ that decreases from $\alpha$ to $\epsilon \varpropto \frac{1}{\alpha}$ as $\widetilde \theta_i(t)$ increases from $\upsilon \varpropto \alpha$ to $\pi/2$.} As $\alpha$ increases, the gap between small angle and large angle is amplified  \textcolor{blue}{(e.g., $f(\widetilde \theta_i(t))$ changes within a relatively large range $[\alpha, \epsilon]$ as $\widetilde \theta_i(t)$ increases within range $[\alpha, \pi/2]$ )}, so is the difference of contribution from those nodes. \textcolor{blue}{However, keeping increasing $\alpha$ is not consistently effective to distinguish the difference of contributions from nodes. Since $\upsilon$ is proportional to $\alpha$, a large $\alpha$ narrows the boundary $[\upsilon, \frac{\pi}{2}]$ where the node contribution should be considered, making the contribution of nodes whose angle lays between $[0, \upsilon]$ indistinguishable. The choice of $\alpha$ is empirically verified in Section \ref{V-C}.}

\subsubsection{Weighting}
\label{IV-C-2}
After getting the contribution mapped using the smoothed angle from each node, we use \textit{Softmax function} to finally calculate the weight of participating nodes for global model aggregation as follows: 
\begin{align*}
\label{eq14} 
\widetilde \psi_i(t) =
    \begin{cases}
       \frac{e^{f(\widetilde \theta_i(t)) }}{\sum_{i^\prime=1}^{|\mathcal{S}_t|} e^{f(\widetilde \theta_{i^\prime}(t))}} & D_m = D_n, \forall m,n \in \mathcal{S}_t \\          
 \frac{D_i e^{f(\widetilde \theta_{i}(t))}}{\sum_{i^\prime=1}^{|\mathcal{S}_t|} D_{i'} e^{f(\widetilde \theta_{i^\prime}(t))}}  & D_m \neq D_n, \exists m,n \in \mathcal{S}_t
    \end{cases}  
 .    \tag{\textcolor{blue}{11}}
\end{align*}

\textcolor{blue}{
From the first line of (\ref{eq14}), if all the participating nodes have the same size of data samples, the proposed \texttt{FedAdp} algorithm will assign weight solely based on their contribution quantified by $e^{f(\widetilde \theta_i(t)) }$. From the 2nd line of (\ref{eq14}), \texttt{FedAdp} will assign weight based on both the contribution and the data size.
\begin{rem}
\label{rem3}
\textcolor{blue}{\rm Different from \texttt{FedAvg}, where the weight for aggregation is solely proportioanl to the size of local datasets (e.g. $\psi_i = D_i/ \sum_{i^\prime=1}^{|\mathcal{S}_t|} D_{i^\prime}$), \texttt{FedAdp}
takes both the data size and the node contribution into consideration when assigning  weights for model aggregation.}
\end{rem}
}
The reason for adopting the Softmax function is twofold: i) 
The output of the Softmax function is a \textit{normalized value} with a larger angle corresponding to a \textcolor{blue}{smaller} weight. ii) Using the Softmax function, each node's contribution can be reinforced or suppressed, depending on the smoothed angle between its gradient and the global gradient. 

The complete procedures of the proposed \texttt{FedAdp} algorithm  are presented in Algorithm \ref{algorithm2} and \texttt{FedAdp} with adaptive weighting strategy leads to the following theorem.
\begin{customthm}{2}
\label{thm2}
\texttt{FedAdp} with weight design $\widetilde \psi_i$ achieves a tighter bound on FL loss decrease in Theorem \ref{thm1} than \texttt{FedAvg} with weight $\psi_i$. 
\end{customthm}
The detailed proof of Theorem \ref{thm2} is presented in Appendix\ref{apex2}.

Compared to \texttt{FedAvg}, \texttt{FedAdp} adopts a simple yet effective strategy that measures the node contribution by quantifying the correlation between the local gradient and the global gradient. Weight for the global model updates can be adaptively assigned based on node contribution rather than evenly averaging, which results in greater FL loss reduction in each global round and accelerates model convergence consequently, as confirmed by our experimental results. 

\begin{algorithm}[t]
 \caption{\texttt{Fed}erated \texttt{Ad}a\texttt{p}tive Weighting \rm (\texttt{FedAdp})}
 \label{algorithm2}
 \begin{algorithmic}[1]
 \renewcommand{\algorithmicrequire}{\textbf{procedure} \	\textsc{Federated Optimization} }
 \renewcommand{\algorithmicensure}{\textbf{Input:} node set $\mathcal{S}, E,B,T, \eta$,  }
 \REQUIRE 
 \ENSURE 
 \STATE Server initializes global model $\mathbf{w}(0)$, global update $ \Delta (0)$, \textcolor{blue}{smoothed angle \textcolor{blue}{$ \widetilde \theta_i(0), i \in \mathcal{S}$}}
 
  \STATE  \hspace*{1em} \textbf{for} $t = \textcolor{blue}{1}, \cdots, T-1$ \textbf{do} \\ 
   \STATE  \hspace*{2em} \textbf{for} node $i \in \mathcal{S}_t$ in parallel \textbf{do} 
  \STATE  \hspace*{3em} $\Delta_i(t) \leftarrow$ \textsc{Local Update} ( $i, \mathbf{w}_i(t-1)$)

     \STATE \hspace*{2em} $\mathbf{w}(t) \leftarrow$ \textsc{Global Update} \\ $\hspace{80pt}$($\Delta_1(t)$ $\Delta_2(t), \cdots, \Delta_{|\mathcal{S}_t|}(t)$ )

  \renewcommand{\algorithmicrequire}{\textbf{procedure} \	\textsc{Local Update}}
 \renewcommand{\algorithmicensure}{\textbf{Input:} node index $ i$, model $\mathbf{w}_i(t-1)$  }
 
 \REQUIRE 
 \ENSURE
 
  \STATE Calculate local updates for $\tau = D_i \frac{E}{\bar B}$ times of SGD with step-size $\eta$ on $F_i( \mathbf{w} )$ \textcolor{blue}{and obtain $\mathbf{w}_i(t)$} using (\ref{eq3})
  \STATE Calculate the model difference $\Delta_i(t) = \mathbf{w}_i (t)	- \mathbf{w} (t-1) $
  \STATE \textbf{return} $\Delta_i(t)$
  
    \renewcommand{\algorithmicrequire}{\textbf{procedure} \	\textsc{Global Update}}
 \renewcommand{\algorithmicensure}{\textbf{Input:} local update $\Delta_1(t), \Delta_2(t), \cdots, \Delta_{|\mathcal{S}_t|}(t)$  }
 \REQUIRE 
 \ENSURE
  \STATE \textcolor{blue}{Calculate the global gradient $ \nabla  F( \mathbf{w} (t)) =  \sum_{i=1}^{|\mathcal{S}_t|} (D_i / \sum_{i^\prime=1}^{|\mathcal{S}_t|} D_{i^\prime}) \nabla  F_i( \mathbf{w}(t)) $, where $\nabla  F_i( \mathbf{w}(t)) = -\Delta_i(t)/\eta $}
  \STATE Calculate instantaneous angle $\theta_i(t)$ by (\ref{eq8})\\
  \STATE Update smoothed angle \textcolor{blue}{$ \widetilde \theta_i(t)$} by (\ref{eq9}) 
   \STATE Calculate weight for model aggregation by (\ref{eq13}), (\ref{eq14})
   
  \STATE Update global model $\mathbb{E}_{i, t} \left[ \widetilde \psi_i(t) \mathbf{w}_i(t-1) \right]$
  
  \STATE \textbf{return} $\mathbf{w}(t)$
  
 \end{algorithmic} 
 \end{algorithm}
 
\section{Evaluation and analysis}
\label{V}
To evaluate the performance of our proposed adaptive weighting algorithm, we implemented \texttt{FedAdp} with PyTorch framework and PySyft library, and studied the image classification task. We evaluated \texttt{FedAdp} by training typical convex and non-convex learning models on two datasets: MNIST and FashionMNIST. Similar to the experiment in section \ref{III-B}, when the different degree of skewness of non-IID dataset is presented, we first investigated how \texttt{FedAdp} outperforms \texttt{FedAvg}\cite{McMahan2017CommunicationEfficientLO} by assigning adaptive weight for model aggregation. Note that our proposed algorithm is not limited by the presence of the IID dataset and can be applied to a general scenario with data heterogeneity as verified in Section \ref{IV-A}. Then, the choice of $\alpha$ for non-linear mapping in \texttt{FedAdp} is discussed in Section \ref{IV-B}. Finally, by tracking the divergence of gradients on participating nodes, we showed \texttt{FedAdp} alleviates the impact brought by the data heterogeneity, compared to \texttt{FedAvg}, which is beneficial to reducing the FL loss in each round and accelerating FL model convergence as discussed in Section \ref{IV-C}. 

We briefly describe our experiment settings as follows.

\begin{figure*}[h]
\centerline{\includegraphics[scale=0.35]{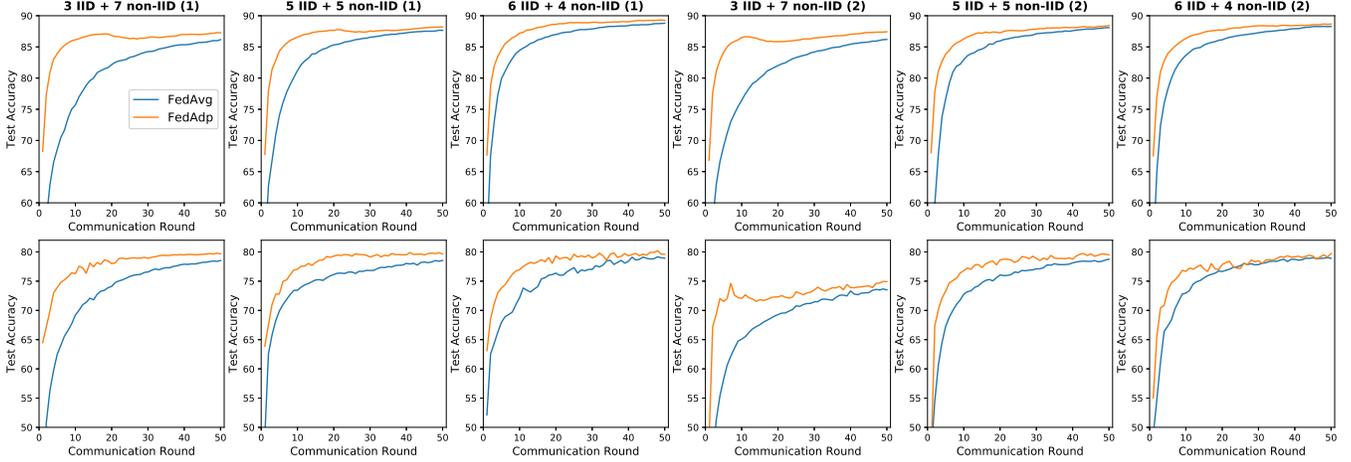}}
\caption{Test accuracy over communication rounds of \texttt{FedAdp} and \texttt{FedAvg} with heterogeneous data distribution over participating nodes using MLR model. Upper and lower subplots correspond to training performance on MNIST and FashionMNIST datasets, respectively.}
\label{fig3}
\end{figure*}

We consider Multinomial Logistic Regression\footnote{For MLR model, the input is a flattened 784-dimensiona (28$\times$28) image, and the output is a class label between 0 and 9. Note that MLR model can be easily extended to strongly-convex setting by adding regularlization term \cite{dinh2019federated}.} (MLR) model and CNN model\footnote{The CNN has 7 layers with the following structure: $\rm 5 \times 5 \times 32$ Convolutional $\rightarrow 2 \times 2$ MaxPool $\rightarrow $ $\rm 5 \times 5 \times 64$ Convolutional $\rightarrow 2 \times 2$ MaxPool $\rightarrow 1024\times512 $ Fully connected $\rightarrow 512\times10 $ Fully connected $\rightarrow$ Softmax (1,663,370 total parameters). All Convolutional and Fully connected layers  are mapped by ReLu activation. The configuration is similar to \cite{McMahan2017CommunicationEfficientLO}.} to represent convex and non-convex learning objective, respectively. we use the number of communication rounds for the FL model to reach a target testing accuracy as a performance metric. Unless otherwise specified, the target accuracy is set to 95\% for training on MNIST, and 80\% for training on FashionMNIST. The number of participating nodes $\mathcal{|S|}=10$, $D_i=600$, $\bar B=50$ for MLR and $\bar B=32$ for CNN, $E=1$, $T=300$, $\eta=0.01$, decay rate $=0.995$, the constant in non-linear mapping function $\alpha=5$. The skewness of the dataset is measured by \textit{$x$-class non-IID}. The dataset for nodes is generated in the same way as in section \ref{III-B}.

\subsection{Data Heterogeneity}
\label{V-A}
We investigate the different number of non-IID nodes with different skewness levels of non-IID data to testify the efficiency of \texttt{FedAdp}. For non-IID data, two skewness cases that $x=1, 2$ are considered. We plot the
test accuracy vs. the communication rounds of federated
learning in Fig. \ref{fig3} and Fig. \ref{fig4} when MLR and CNN models are adopted, respectively.
\subsubsection{MLR Model}
Given the learning capability of MLR is limited, instead of setting a target accuracy, we simply train a model over 50 global rounds. We plot the test accuracy vs. the communication rounds of federated learning algorithms in Fig. \ref{fig3}. From Fig. \ref{fig3}, we can tell \texttt{FedAdp} always outperforms \texttt{FedAvg} when the nodes with non-IID dataset are present. In addition, \texttt{FedAdp} converges very fast in the early training stage, and the superiority of \texttt{FedAdp} is more prominent when the proportion of nodes with non-IID datasets is larger. It is noted that the gap between \texttt{FedAdp} and \texttt{FedAvg} over 50 global rounds is not conspicuous because of the simplicity of the MLR model. Different weighting strategies will not make much difference when the model is reaching its learning capability. In contrast, the weighting strategy will consistently impact the FL training process when a more complex neural network model is applied, as shown in the following experiment.

\subsubsection{CNN Model}
\begin{figure*}[t]
\centerline{\includegraphics[scale=0.35]{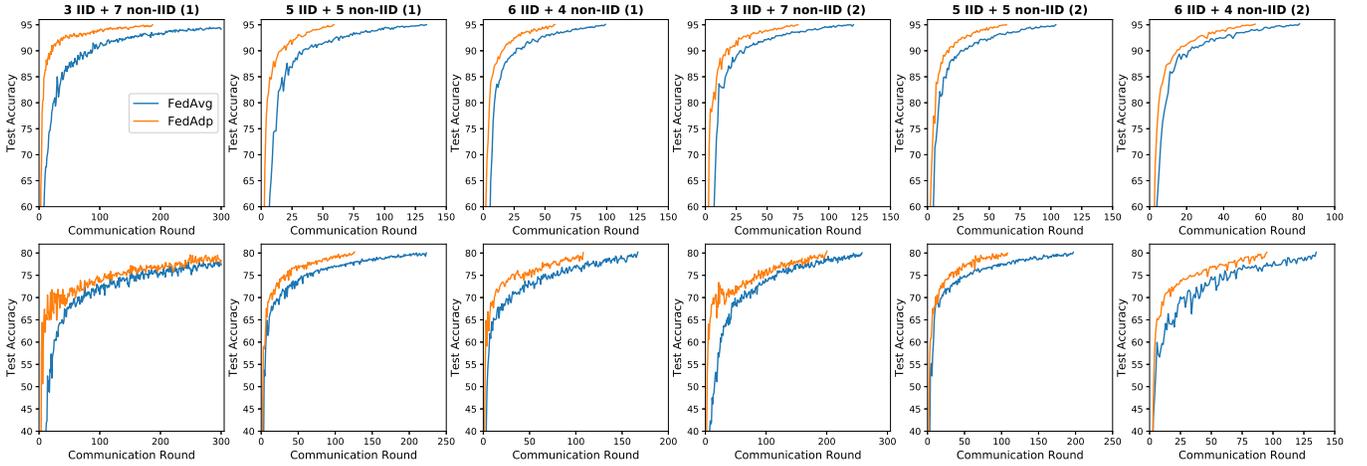}}
\caption{Test accuracy over communication rounds of \texttt{FedAdp} and \texttt{FedAvg} with heterogeneous data distribution over participating nodes using CNN model. Upper and lower subplots correspond to training performance on MNIST and FashionMNIST datasets, respectively.}
\label{fig4}
\end{figure*}

We plot the test accuracy vs. the communication rounds of federated
learning in Fig. \ref{fig4}. From Fig. \ref{fig4}, we can tell \texttt{FedAdp} always outperforms \texttt{FedAvg} when the nodes with non-IID dataset are present. In particular, \texttt{FedAdp} converges very fast in the early training stage since the gradient divergence is more obvious in the initial rounds, which makes the effect of assigning adaptive weight for updating the global model even more significant.
\begin{table}[h]
\centering
\renewcommand{\arraystretch}{1.3}
\caption{Number of communication rounds to reach a target
accuracy for \texttt{FedAdp}, versus  \texttt{FedAvg}\cite{McMahan2017CommunicationEfficientLO}, within 300 rounds. N/A refers that algorithms can not reach target accuracy before termination where the highest test accuracy is shown}
\label{table1}
{\scriptsize
\begin{tabular}{c c c c c}
    \hline
  \multicolumn{4}{c}{ \textsc{ \textbf{ MNIST 95\% Accuracy}} } \\
    \multicolumn{4}{c}{ \textsc{ 1-class non-IID} }\\
    
& 3 IID + 7 non-IID & 5 IID + 5 non-IID & 6 IID + 4 non-IID\\
    \texttt{FedAvg} &   N/A (94.48\%) &   133 &  99\\
    \texttt{FedAdp}    & \textbf{187}    &   \textbf{61} & \textbf{58}   \\  
     \multicolumn{4}{c}{ \textsc{ 2-class non-IID} }\\   
    \texttt{FedAvg}  & 120  &  104 & 81 \\
    \texttt{FedAdp}   & \textbf{75}  & \textbf{59}   & \textbf{52} \\
    \hline
    
    \multicolumn{4}{c}{ \textsc{ \textbf{ Fashion MNIST 80\% Accuracy}} } \\
    \multicolumn{4}{c}{ \textsc{ 1-class non-IID} }\\
    
& 3 IID + 7 non-IID & 5 IID + 5 non-IID & 6 IID + 4 non-IID\\
    \texttt{FedAvg} &  N/A (77.31\%) & 222 & 167  \\
      \texttt{FedAdp} & N/A (\textbf{79.5}\%)  & \textbf{125} & \textbf{107}  \\
     \multicolumn{4}{c}{ \textsc{ 2-class non-IID} }\\
    \texttt{FedAvg} &  258  & 196  &   134\\
    \texttt{FedAdp} & \textbf{207}    & \textbf{107} &   \textbf{94}\\
    \hline
\end{tabular}
}
\end{table}

To measure the effectiveness of \texttt{FedAdp}, we count the number of communication rounds needed to reach a target accuracy when \texttt{FedAdp} is adopted.
Each entry in Table \ref{table1} shows the number of communication rounds necessary to achieve a test accuracy of 95\% for CNN on MNIST and 80\% for FashionMNIST. The bold number indicates the better result achieved by \texttt{FedAdp}, as compared to \texttt{FedAvg}. \texttt{FedAdp} decreases the number of communication rounds by up to 54.1\% and 43.2\% for the MNIST task when non-IID nodes are at 1-class and 2-class non-IID setting, respectively. For the FashionMNIST task, the corresponding decreases are up to 43.7\% and 45.4\%, respectively. In the cases when the target accuracy is not reachable before 300 rounds, \texttt{FedAdp} always terminates with higher testing accuracy.

Previously, two extremely skewness cases that $x=1, 2$ are considered, while the superiority of the proposed weighting strategy is not limited to extreme cases. To verify the proposed weighting strategy in a more general data heterogeneity case, we consider the CNN model for the MNIST dataset in the following two cases.   
\begin{itemize}
\item \textit{Case 1}: The number of classes of data samples owned by node $i$, denoted by $x_i$, is randomly selected from the set $\{1, 2, \cdots, 10\}$ without overlapping. Whereafter, the data samples on each node are randomly selected from the $x_i$-subset of the training dataset.
\item \textit{Case 2}: 
For half of the nodes, their $x_i$ (i.e., the number of classes of data samples) is selected following the uniform distribution $\mathcal{U}(1, 5)$, whereas for the other half, $x_i$ follows the uniform distribution $\mathcal{U}(6, 10)$. The data samples on each node are randomly selected from the $x_i$-subset of the training dataset.
\end{itemize}

\begin{figure}[h]
  \centerline{\includegraphics[scale=0.34]{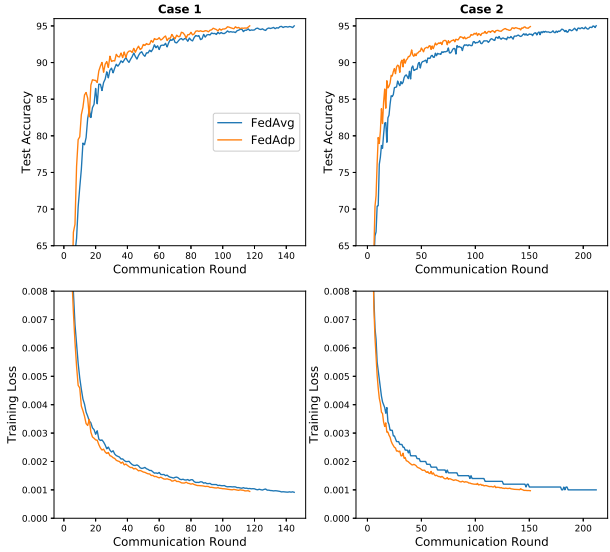}}
  \caption{FL training performance over communication rounds when \texttt{FedAdp} is adopted considering general heterogeneous data distribution over participating nodes. The top row and bottom row represent the test accuracy and \textcolor{blue}{training loss} over the communication round, respectively.}
  \label{fig7}
\end{figure}

From Fig. \ref{fig7}, we can see \texttt{FedAdp} outperforms \texttt{FedAvg} in both cases. In both cases, the convergence performance is worse than the result in Fig. \ref{fig4} because the number of IID nodes is small and the local dissimilarity is greater in these two cases. However, it is clear by measuring node contribution, \texttt{FedAdp} is more rapid in reducing FL loss in each global round thus accelerating model convergence, even without the participation of IID nodes.  
\subsection{Choosing $\alpha$}
\label{V-C}
One natural question is how to determine $\alpha$ for non-linear function. A large $\alpha$ may increase the convergence by emphasizing the difference of contribution from participating nodes, which hastens model convergence in the initial training stage. Meanwhile, since $\upsilon$ is proportional to $\alpha$, a large $\alpha$ also narrows the boundary $[\upsilon, \frac{\pi}{2}]$ where the node contribution should be considered, making the contribution of nodes whose angle lays between $[0, \upsilon]$ indistinguishable.

\begin{figure}[h]
 \centerline{\includegraphics[scale=0.55]{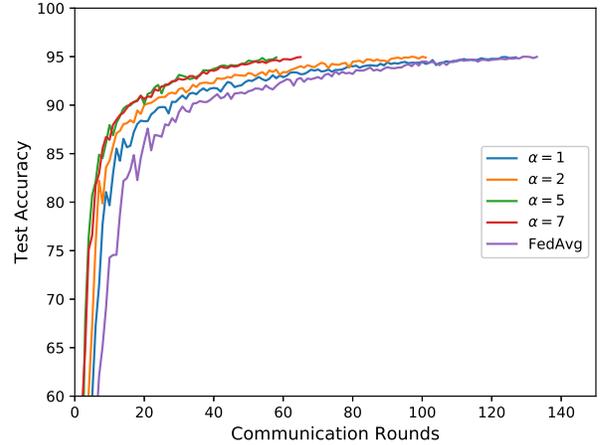}}
  \caption{Effect of setting $\alpha$ on federated learning performance. Data heterogeneity setting is 5 IID + 5 non-IID (1) and CNN model is adopted.}
  \label{fig5}
\end{figure}

We heuristically choose $\alpha \in \mathbb{Z}^{+}$ in the ascending order. From Fig. \ref{fig5}, increasing $\alpha$ leads to faster convergence since the gap between small angle and large angle is amplified, so is the difference of contribution from those nodes. However, a larger $\alpha$ is not always effective, especially after the initial training stage. Empirically, the best $\alpha$ is 5 for our experimental setting. 

\subsection{Divergence Measurement}
\label{V-D}
\begin{figure}[h]
  \centerline{\includegraphics[scale=0.34]{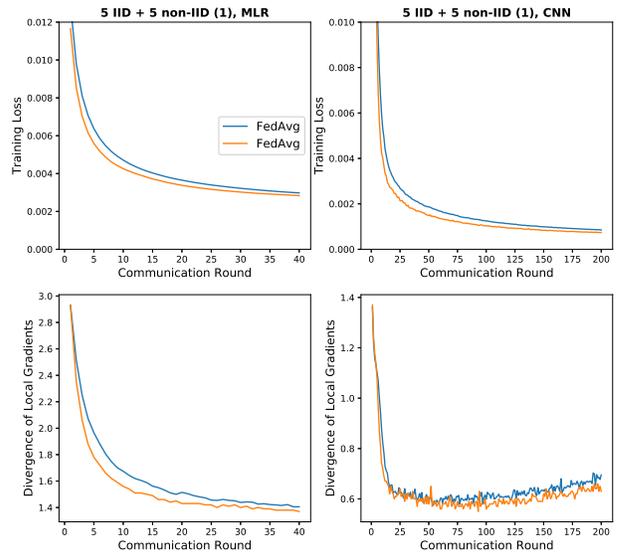}}
  \caption{The connection between the model test loss and the divergence across local gradients. The proposed weighting strategy \texttt{FedAdp} gives an impact on alleviating the divergence brought by nodes with skewed datasets. (1) Top row: the \textcolor{blue}{training loss} on the MNIST dataset under one data heterogeneity setting (5 IID + 5 non-IID (1)). (2) Bottom row: the corresponding divergence measurement.}
  \label{fig6}
\end{figure}

Finally, in Fig. \ref{fig6}, we take one experimental case as an example to demonstrate the divergence of local gradients, which captures the overall data heterogeneity of participating nodes. In particular, we track the divergence of gradients over all participating nodes, which is measured by $\sum_i^{\mathcal{S}_t} \frac{1}{|\mathcal{S}_t|} \Vert \nabla F( \mathbf{w}) -\nabla F_i(\mathbf{w}) \Vert$. Empirically, we observe that our proposed weighting strategy leads to smaller divergence among participating nodes, and the smaller the divergence, the smaller the FL loss. As $\mathbf{w}(t)$ is not a stationary solution along with the training, aggregation by \texttt{FedAdp} is seen as a regularization process that restrains the local weight $\mathbf{w}_i(t+1)$ trained by skewed datasets from being deviatory, which lowers the model divergence and consequently accelerates the convergence.

\section{Conclusion}
\label{VI}
In this paper, we have presented our design of \texttt{FedAdp} algorithm that assigns nodes with different weights for updating the global model in each round adaptively to reduce the communication rounds of FL training in the presence of non-IID data. We argue that non-IID data exacerbates the model divergence and observe the nodes with non-IID data make a smaller (or even negative) contribution to the global model aggregation than the nodes with IID data. We have proposed to measure the node contribution based on the angle between local gradient and global gradient and designed a non-linear mapping function to quantify node contribution. We have designed an adaptive weighting strategy that assigns weight proportional to node contribution instead of according to the size of local datasets. The simple yet effective strategy is able to reinforce positive (suppress negative) node contribution dynamically, leading to a significant communication round reduction. Its performance superiority over \texttt{FedAvg} is verified both
theoretically and experimentally. We have shown that FL training with \texttt{FedAdp} has reduced the communication rounds by up to 54.1\% on the MNIST dataset and up to 45.4\% on the FashionMNIST dataset compared to \texttt{FedAvg}.


%


\ifCLASSOPTIONcaptionsoff
  \newpage
\fi



\bibliographystyle{ieeetr}
\bibliography{ref} 

\vspace{20pt}
\appendices
\section*{appendix}
\subsection{Proof of Theorem \ref{thm1}}
\label{apex1}
From the $\beta$-Lipschitz smoothness of $F(\mathbf{w})$ in Lemma \ref{lem1} and Taylor expansion, we have 
\begin{align*}
\label{a1}
F(\mathbf{w}(t+1)) \leq & F( \mathbf{w}(t)) + \langle \nabla F( \mathbf{w}(t)),  \mathbf{w}(t+1) - \mathbf{w}(t) \rangle \\
& + \frac{\beta}{2}  \Vert\mathbf{w}(t+1) - \mathbf{w}(t) \Vert^2. \tag{A1}
\end{align*}

The last two terms on the right hand side of the  \textcolor{blue}{above} inequality are bounded respectively as: 

$\bullet$ \textit{Bounding} $\Vert\mathbf{w}(t+1) - \mathbf{w}(t) \Vert^2$:
By the definition of the global aggregation for $\mathbf{w}(t+1)$, we have
\begin{align*}
\label{a2}
 \Vert \mathbf{w}(t+1) - \mathbf{w}(t)\Vert 
 & =  \mathbb{E}_{i|t} \left[  \Vert \mathbf{w}_i(t+1) - \mathbf{w}(t)\Vert \right].
 \tag{A2}
\end{align*}

By following SGD optimization, for each term within the expectation in the right hand side of \ref{a2}, we have
\begin{align*}
\label{a3}
 \mathbf{w}_i (t+1) & =  \mathbf{w} (t) - \eta  \nabla F_i( \mathbf{w}(t)).
 \tag{A3}
\end{align*}

Therefore, 
\begin{align*}
\label{a4}
 \Vert \mathbf{w}(t+1) - \mathbf{w}(t)\Vert^2 
 & =  (\mathbb{E}_{i|t} \left[  \Vert \mathbf{w}_i(t+1) - \mathbf{w}(t)\Vert \right])^2 \\
 & = \eta^2  (\mathbb{E}_{i|t} \left[ \Vert  \nabla F_i( \mathbf{w}(t)) \Vert  \right])^2\\
  &\stackrel{1}\leq  \eta^2  \mathbb{E}_{i|t} \left[ \Vert  \nabla F_i( \mathbf{w}(t)) \Vert^2 \right],  
 \tag{A4}
\end{align*}
where inequality 1 holds because of Cauchy-Schwarz inequality.

$\bullet$ \textit{Bounding} $\langle \nabla F( \mathbf{w}(t)),  \mathbf{w}(t+1) - \mathbf{w}(t) \rangle$: 
Again, by the definition of the global aggregation for $\mathbf{w}(t+1)$ and \ref{a3} we have
\begin{align*}
\label{a5}
 & \langle \nabla F( \mathbf{w}(t)),  \mathbf{w}(t+1) - \mathbf{w}(t) \rangle \\ 
 = & -\eta \mathbb{E}_{i|t}\left[     \langle \nabla F( \mathbf{w}(t)),  \nabla F_i( \mathbf{w}(t))  \rangle \right].  \tag{A5}
\end{align*}

The expectation term in \ref{a5} can be further rewritten as 
\begin{align*}
\label{a6}
& \mathbb{E}_{i|t}\left[     \langle \nabla F( \mathbf{w}(t)),  \nabla F_i( \mathbf{w}(t))  \rangle \right] \\
 = & \mathbb{E}_{i|t} \left[   \frac{\langle \nabla  F( \mathbf{w}(t)), \nabla  F_i( \mathbf{w}(t))\rangle}{\Vert \nabla  F( \mathbf{w}(t))\Vert  \Vert \nabla  F_i( \mathbf{w}(t))\Vert} \cdot
 \Vert \nabla  F( \mathbf{w}(t))\Vert  \Vert \nabla  F_i( \mathbf{w}(t))\Vert \right] \\
\stackrel{2}\geq & \mathbb{E}_{i|t} \left[   \frac{\langle \nabla  F( \mathbf{w}(t)), \nabla  F_i( \mathbf{w}(t))\rangle}{\Vert \nabla  F( \mathbf{w}(t))\Vert  \Vert \nabla  F_i( \mathbf{w}(t))\Vert} \cdot  \frac{\Vert F_i( \mathbf{w}(t))\Vert^2}{B} \right], \tag{A6}
\end{align*}
where inequality 2 comes from Assumptions \ref{ass2} that local dissimilarity is upper bounded by $B$.

Plugging \ref{a6} into \ref{a5}, then the last two terms on the right hand side of \ref{a1} are expressed as
\begin{align*}
\label{a7}
& \langle \nabla F( \mathbf{w}(t)),  \mathbf{w}(t+1) - \mathbf{w}(t) \rangle + \frac{\beta}{2}  \Vert\mathbf{w}(t+1) - \mathbf{w}(t) \Vert^2 \\
& \leq -\eta \mathbb{E}_{i|t} \left[   \frac{\langle \nabla  F( \mathbf{w}(t)), \nabla  F_i( \mathbf{w}(t))\rangle}{\Vert \nabla  F( \mathbf{w}(t))\Vert  \Vert \nabla  F_i( \mathbf{w}(t))\Vert} \cdot  \frac{\Vert F_i( \mathbf{w}(t))\Vert^2}{B} \right]  
\\ 
 & \quad +  \frac{\beta \eta^2 }{2} \mathbb{E}_{i|t} \left[ \Vert  \nabla F_i( \mathbf{w}(t)) \Vert^2 \right]  \\
\stackrel{3}\leq & -\eta \mathbb{E}_{i|t} \left[ \left(  \frac{\langle \nabla  F( \mathbf{w}(t)), \nabla  F_i( \mathbf{w}(t))\rangle}{\Vert \nabla  F( \mathbf{w}(t))\Vert  \Vert \nabla  F_i( \mathbf{w}(t))\Vert} - \frac{B \beta\eta}{2} \right) \cdot \frac{A^2}{B} \Vert \nabla F( \mathbf{w}) \Vert ^2 \right],
 \tag{A7}
\end{align*}
where inequality 3 holds because of Assumptions \ref{ass2} that local dissimilarity is lower bounded by $A$.

Finally, Theorem \ref{thm1} is proved by substituting \ref{a7} into \ref{a1}.
\subsection{Proof of Theorem \ref{thm2}}
\label{apex2}
\textcolor{blue}{We consider the general case that participating nodes have a different number of data samples. For node $i$ with data size $D_i$, we create $D_i$ virtual nodes, each with a unit sample size. Hereinafter, we use index $(i,j), j \in \{1, \cdots, D_i\}$ to denote the $j$-th virtual node split from the participating node $i, i \in \mathcal{S}_t$, where the gradient information is kept on virtual nodes as on the participating node (e.g., $\nabla F_{i,j}( \mathbf{w}(t) = \nabla F_{i}( \mathbf{w}(t)), \theta_{i,j} = \theta_i$). As such, all virtual nodes split by node $i$ share the same weight (i.e., $\widetilde \psi_{i,j}(t) = \widetilde \psi_{i,k}(t), \forall j,k \in \{1, \cdots, D_i\}$), where $\widetilde \psi_{i,j}(t)$ denotes the weight for virtual node $(i,j)$. The weight of node $i$ is $\widetilde \psi_{i}(t) = \sum_{j=1}^{D_i}\widetilde \psi_{i,j}(t) = D_i\widetilde \psi_{i,j}(t)$.}

\textcolor{blue}{From (\ref{eq8}), $\theta_{i,j} = \theta_i$ monotonically decreases with $\frac{\langle \nabla  F( \mathbf{w}(t)), \nabla F_{i}( \mathbf{w}(t))\rangle}{\Vert \nabla  F( \mathbf{w}(t))\Vert  \Vert\nabla F_{i}( \mathbf{w}(t))\Vert}$. From (\ref{eq13}), $f(\cdot)$ is a decreasing function of $\theta$. Thus, by that $\widetilde \psi_{i,j} (t) = \frac{e^{f(\widetilde \theta_{i,j}(t))}}{\sum_{i^\prime=1}^{|\mathcal{S}_t|} D_{i'} e^{f(\widetilde \theta_{i^\prime}(t))}}$, we can see $\widetilde \psi_{i,j} (t)$ monotonically increases with $\frac{\langle \nabla  F( \mathbf{w}(t)), \nabla F_i( \mathbf{w}(t))\rangle}{\Vert \nabla  F( \mathbf{w}(t))\Vert  \Vert \nabla F_i( \mathbf{w}(t))\Vert}$. Therefore, generic $\widetilde \psi_{i,j} (t)$ satisfies the following criterion,}
\begin{align*}
\label{eq11}
& \widetilde \psi_{i,j} (t)  \varpropto \frac{\langle \nabla  F( \mathbf{w}(t)), \nabla F_i( \mathbf{w}(t))\rangle}{\Vert \nabla  F( \mathbf{w}(t))\Vert  \Vert \nabla F_i( \mathbf{w}(t))\Vert} 
\\
& \widetilde \psi_{i,j}(t)  \geq 0 \; \; \forall i,j, t
\\
& \sum_{i=1}^{|\mathcal{S}_t|} \sum_{j=1}^{D_i} \widetilde \psi_{i,j}(t) = \sum_{i=1}^{|\mathcal{S}_t|} \widetilde \psi_{i}(t) =1,
   \tag{\textcolor{blue}{B1}}
\end{align*}
with the corresponding bound of the expected loss being
\begin{align*}
\label{12}
& F(\mathbf{w}(t+1)) \leq  F( \mathbf{w}(t))  \\ & -\eta \sum_i^{|\mathcal{S}_t|} \left(  \frac{\langle \nabla  F( \mathbf{w}(t)), \nabla  F_i( \mathbf{w}(t))\rangle}{\Vert \nabla  F( \mathbf{w}(t))\Vert  \Vert \nabla  F_i( \mathbf{w}(t))\Vert} \widetilde \psi_i (t) - \frac{B\beta\eta}{2} \right) \cdot \frac{A^2}{B} \Vert \nabla F( \mathbf{w}) \Vert ^2 . \tag{B2}
\end{align*}
where $\widetilde \psi_i (t)$ is defined as in (\ref{eq14}).

In order to compare the expected loss achieved by \texttt{FedAdp} and \texttt{FedAvg}, one can simply measure the expectation term in (\ref{eq5}). \textcolor{blue}{We use $u_{i,j}$ to denote the contribution from virtual node $j$ of participating node $i$ for model aggregation. In each global round, we sort the contribution from all the virtual nodes that is measured by the correlation $\frac{\langle \nabla  F( \mathbf{w}(t)), \nabla  F_i( \mathbf{w}(t))\rangle}{\Vert \nabla  F( \mathbf{w}(t))\Vert  \Vert \nabla  F_i( \mathbf{w}(t))\Vert}$ between the local gradient and the global gradient in descending order, that is $u_{1,1} =u_{1,2} =\cdots = u_{1,D_1} \geq u_{2,1} =u_{2,2} =\cdots = u_{2,D_2} \geq \cdots \geq u_{|\mathcal{S}_t|,1} =u_{|\mathcal{S}_t|,2} =\cdots = u_{|\mathcal{S}_t|,D_{|\mathcal{S}_t|}}$. Apparently, the weight assigned to virtual node in \texttt{FedAdp} should follow the same order $\widetilde \psi_{1,1} =\widetilde \psi_{1,2} =\cdots = \widetilde \psi_{1,D_1} \geq \widetilde \psi_{2,1} =\widetilde \psi_{2,2} =\cdots = \widetilde \psi_{2,D_2} \geq \cdots \geq \widetilde \psi_{|\mathcal{S}_t|,1} =\widetilde \psi_{|\mathcal{S}_t|,2} =\cdots = \widetilde \psi_{|\mathcal{S}_t|,D_{|\mathcal{S}_t|}}$, with $\sum_i \sum_j\widetilde \psi_{i,j}=1$. 
As such, by Chebyshev's inequality \cite{marshall1960multivariate}, we have the following hold for any $u_{m,j}$, $u_{n,j'}$,}
\begin{align*}
\label{b1}
\bar \psi (u_{m,j} - u_{n,j'})( \frac{\widetilde \psi_{m,j}}{\bar \psi_{m,j}}  - \frac{\widetilde \psi_{n,j'}}{\bar \psi_{n,j'}})   \geq 0 \\
\bar \psi [ u_{m,j}\widetilde \psi_{m,j}\bar \psi_{n,j'} + u_{n,j'}\widetilde \psi_{n,j'}\bar \psi_{m,j} ]  \\ \geq  \bar \psi [u_{m,j}\widetilde \psi_{n,j'}\bar \psi_{m,j} + u_{n,j'}\widetilde \psi_{m,j}\bar \psi_{n,j'}], 
  \tag{\textcolor{blue}{B3}}
\end{align*}
\textcolor{blue}{where $\bar \psi = \bar \psi_{m,j} = \bar \psi_{n,j'} = \frac{1}{D}$ denotes the weight of \texttt{FedAvg} for all virtual nodes with $D=\sum_i^{|\mathcal{S}_t|}D_i$.}

\textcolor{blue}{Adding all the $D^2$ inequalities, we have,}

\begin{align*}
\label{b2}
\begin{split}
& \bar \psi \left[ \sum_{m=1}^{|\mathcal{S}_t|} \sum_{j=1}^{D_m} \sum_{n=1}^{|\mathcal{S}_t|} \sum_{j'=1}^{D_n} 
 u_{m,j}\widetilde \psi_{m,j}\bar \psi_{n,j'} + u_{n,j'}\widetilde \psi_{n,j'}\bar \psi_{m,j} \right] \\  \geq & \bar \psi \left[ \sum_{m=1}^{|\mathcal{S}_t|} \sum_{j=1}^{D_m} \sum_{n=1}^{|\mathcal{S}_t|} \sum_{j'=1}^{D_n}  u_{m,j}\widetilde \psi_{n,j'}\bar \psi_{m,j} + u_{n,j'}\widetilde \psi_{m,j}\bar \psi_{n,j'} \right] 
\end{split}\\
\begin{split}
&\sum_{m=1}^{|\mathcal{S}_t|} \sum_{j=1}^{D_m} u_{m,j}\widetilde \psi_{m,j} \underbrace{\sum_{n=1}^{|\mathcal{S}_t|} \sum_{j'=1}^{D_n}\bar \psi_{n,j'}}_{=1} + \sum_{n=1}^{|\mathcal{S}_t|} \sum_{j'=1}^{D_n} u_{n,j'}\widetilde \psi_{n,j'} \underbrace{\sum_{m=1}^{|\mathcal{S}_t|} \sum_{j=1}^{D_m} \bar \psi_{m,j}}_{=1}  
\\  \geq & \sum_{m=1}^{|\mathcal{S}_t|} \sum_{j=1}^{D_m}u_{m,j}  \bar \psi_{m,j} \underbrace{\sum_{n=1}^{|\mathcal{S}_t|} \sum_{j'=1}^{D_n} \widetilde \psi_{n,j'}}_{=1} + \sum_{n=1}^{|\mathcal{S}_t|} \sum_{j'=1}^{D_n} u_{n,j'}\bar \psi_{n,j'} \underbrace{\sum_{m=1}^{|\mathcal{S}_t|} \sum_{j=1}^{D_m} \widetilde \psi_{m,j}}_{=1}
\end{split}\\
\begin{split}
 2 \cdot \sum_{m=1}^{|\mathcal{S}_t|} \sum_{j=1}^{D_m} u_{m,j}\widetilde \psi_{m,j}   \geq 2 \cdot \sum_{m=1}^{|\mathcal{S}_t|} \sum_{j=1}^{D_m} u_{m,j}\bar \psi_{m,j}\\
\underbrace{\sum_{m}  u_m \widetilde \psi_m}_{\rm FedAdp}   \stackrel{4}\geq \underbrace{\sum_m u_m \psi_m}_{\rm FedAvg}.
\end{split}
 \tag{\textcolor{blue}{B4}}
\end{align*}
\textcolor{blue}{where $u_m = u_{m,1}= \cdots=u_{m,D_m}$. Inequality 4 holds because $\widetilde \psi_{m} = \widetilde \psi_{m,j} \cdot D_m$ and $\psi_m = \bar \psi_{m,j} \cdot D_m$ with $\widetilde \psi_m$ and $\psi_m$ denoting the weight for model aggregation in \texttt{FedAdp} and \texttt{FedAvg}, respectively. The equality 4 holds when $u_i=u_j, \forall i,j \in \mathcal{S}_t$.}
 
Due to the greater expectation term in (\ref{eq5}). \texttt{FedAdp} results in greater decrease of FL loss in each global round, as compared to \texttt{FedAvg}. This completes the proof.

\end{document}